%% file: main_arxiv.tex
\documentclass[twoside]{article}

\usepackage[accepted]{aistats_arxiv_2026}
\input{_imports.tex}

\begin{document}

\runningtitle{Subgroup Validity in ML for Echocardiogram Data}

\twocolumn[
\aistatstitle{Subgroup Validity in Machine Learning for Echocardiogram Data}
\aistatsauthor{
Cynthia Feeney\footnotemark[1], Shane Williams \footnotemark[1], Benjamin S. Wessler\footnotemark[2],
and Michael C. Hughes\footnotemark[1]}

\aistatsaddress{\footnotemark[1]Department of Computer Science, Tufts University, Medford, MA, USA
\\
\footnotemark[2]Division of Cardiology, Tufts Medical Center, Boston, MA, USA}

]

\begin{abstract}
    \input{abstract.tex}
\end{abstract}
\input{sections/introduction}

\input{sections/related_work}

\input{sections/data}

\input{sections/analysis}

\input{sections/experiments}

\input{sections/conclusion}


\paragraph*{Data and Code Availability}
We utilize the TMED-2 \citep{huang2022TMED2} and MIMIC-IV-ECHO \citep{mimicivecho} datasets throughout this paper.
These datasets are available after completing a user credentialing process at \citet{tmed2web} and \citet{mimicivecho} respectively.
Code is provided at \url{https://github.com/tufts-ml/AS-classifier-subgroup-eval}.

\paragraph*{Institutional Review Board (IRB).}
Our use of non-public deidentified data was approved by our local IRB when necessary.
Parts of this study dealing with deidentified third-party datasets already available to researchers worldwide did not require IRB approval.

\bibliographystyle{abbrvnat}
\bibliography{ml4h2025.bib,refs_from_zotero.bib}

\appendix
\onecolumn 

\counterwithin{table}{section}
\setcounter{table}{0}
\counterwithin{figure}{section}
\setcounter{figure}{0}

\input{sections/appendix}

\end{document}

%% file: _imports.tex
\usepackage[utf8]{inputenc} 
\usepackage[T1]{fontenc} 
\usepackage{microtype} 
\usepackage{nicefrac} 

\usepackage{enumitem}
\setlist[itemize]{noitemsep, topsep=0pt}
\setlist[enumerate]{noitemsep, topsep=0pt}

\usepackage[round]{natbib}

\usepackage{titletoc}
\usepackage[page,header]{appendix}

\usepackage{url} 
\usepackage{hyperref}
\hypersetup{
    colorlinks,
    citecolor=blue,
    filecolor=blue,
    linkcolor=blue,
    urlcolor=blue,
}

\usepackage{graphicx}
\usepackage{chngcntr} 
\begingroup
    \expandafter\ifx\csname pdfsuppresswarningpagegroup\endcsname\relax\else\global\pdfsuppresswarningpagegroup=1\relax\fi
\endgroup
\usepackage[inkscapelatex=false]{svg} 
\usepackage{tikz}
\makeatletter
\let\Ginclude@graphics\@org@Ginclude@graphics
\makeatother

\usepackage{makecell}
\usepackage{tabularx}
\usepackage{booktabs}
\usepackage{multirow}
\usepackage{colortbl}
\usepackage{tablefootnote}
\usepackage{array}

\usepackage{algpseudocode}

\usepackage{amsmath,amssymb,amsfonts,amsthm}

\usepackage{caption}
\usepackage{subcaption}
\usepackage[switch]{lineno}

%% file: abstract.tex
Echocardiogram datasets enable training deep learning models to automate interpretation of cardiac ultrasound, thereby expanding access to accurate readings of diagnostically-useful images.
However, the gender, sex, race, and ethnicity of the patients in these datasets are underreported and subgroup-specific predictive performance is unevaluated.
These reporting deficiencies raise concerns about subgroup validity that must be studied and addressed before model deployment.
In this paper, we show that current open echocardiogram datasets are unable to assuage subgroup validity concerns.
We improve sociodemographic reporting for two datasets: TMED-2 and MIMIC-IV-ECHO. 
Analysis of six open datasets reveals no consideration of gender-diverse patients and insufficient patient counts for many racial and ethnic groups. 
We further perform an exploratory subgroup analysis of two published aortic stenosis detection models on TMED-2.
We find insufficient evidence for subgroup validity for sex, racial, and ethnic subgroups.
Our findings highlight that more data for underrepresented subgroups, improved demographic reporting, and subgroup-focused analyses are needed to prove subgroup validity in future work.

%% file: sections/introduction.tex
\section{Introduction}
\label{sec:intro}

\input{tables/new_data}

Ultrasound imaging of the human heart, also known as echocardiography, shows rich information about cardiac structure and function.
In clinical practice, transthoracic echocardiograms (TTEs) provide non-invasive evidence for heart health measurements, disease diagnosis, and treatment planning.
To improve reliability and expand access to care, recent research has pushed for automated interpretation of echocardiograms by computer vision systems~\citep{zhangFullyAutomatedEchocardiogram2018,wehbeDeepLearningCardiovascular2023,myhreArtificialIntelligenceenhancedEchocardiography2025}. 

Training automated systems requires the availability of labeled datasets, in which each scan is associated with a desired measurement or diagnostic label provided by a human expert.
In the last decade, several prominent datasets have been released openly to the research community, including efforts from European hospitals such as Unity \citep{unity} and CAMUS~\citep{camus} as well as efforts from the United States such as EchoNet~\citep{echonet-dyn,echonet-lvh},  TMED-2~\citep{huang2022TMED2}, and MIMIC-IV-ECHO~\citep{mimicivecho}.
These open datasets have catalyzed clinically-focused work on improved heart health measurements~\citep{unity,zhangDevelopmentAutomatedNeural2024}, disease detection~\citep{ahmadiTransformerBasedSpatioTemporalAnalysis2024,karmiyMachineLearningenabledScreening2025}, and multi-task models~\citep{holsteCompleteAIEnabledEchocardiography2025}.
Open TTE data have also helped ML methodologists benchmark cross-site representation learning~\citep{alaaETABBenchmarkSuite2022} and methods for limited labeled data~\citep{huangSystematicComparisonSemisupervised2024}.

In this paper, we use open TTE datasets to investigate \emph{subgroup validity}: evidence that a model is suitable for use in subgroups of interest. 
Classically underrepresented patient subgroups have disproportionate cardiovascular disease morbidity and mortality, with documented differences across sex \citep{havranek2015social, porter2024reporting}, race \citep{havranek2015social, south_asian_rep}, sexuality \citep{caceres2017systematic}, and gender \citep{streed2017cardiovascular, porter2024reporting}.
These disparities are likely to worsen because of algorithmic bias, where a machine learning model deployed in a care pathway biases care for a subgroup \citep{paulusPredictablyUnequalUnderstanding2020}.
Therefore subgroup analyses, experiments that can provide evidence of subgroup validity, are crucial for ensuring that automated TTE interpretation improves the health of all patients.


Better reporting of sociodemographic data is the primary prerequisite for subgroup analyses, as patient subgroup membership is necessary to calculate subgroup performance.
The TRIPOD best practice checklist has called for subgroup reporting for the last decade~\citep{collinsTransparentReportingMultivariable2015,collins2024tripod+ai}.
Unfortunately, Table \ref{tab:new_data} shows that most existing open TTE datasets lack adequate reporting of subgroup composition.

Previous work has emphasized that understanding how predictive models perform across subgroups should be a key part of model evaluation \citep{rajkomarEnsuringFairnessMachine2018,leismanDevelopmentReportingPrediction2020,gichoyaEquityEssenceCall2021}. 
Reporting performance for key subgroups is now item 23a of the TRIPOD-AI checklist \citep{collins2024tripod+ai}, though it is a recent addition \citep{gichoyaEquityEssenceCall2021}.
Understandably, widespread adoption lags behind the prescribed best practices, with Table \ref{tab:new_data} showing no previous reporting of model performance across subgroups.

For our subgroup analysis, we focus on predictive models for aortic stenosis (AS). 
Previous evaluations of predictive models for AS focus on aggregate performance on a test set randomly sampled from the overall patient population~\citep{ahmadiTransformerBasedSpatioTemporalAnalysis2024,huangSemiSupervisedMultimodalMultiInstance2025}.
This common ``one-size-fits-all'' evaluation places more weight on subgroups that are more frequent and obscures potentially disproportionate performance across subgroups.
Even if a model performs well in aggregate, a large body of work reminds us that ``generalization across groups should not be
assumed'' \citep{paulusPredictablyUnequalUnderstanding2020}.

To work towards evidence of subgroup validity for AS detection and other tasks supported by open TTE data, our main contributions are:
\begin{enumerate}[leftmargin=*]
\item New and improved sociodemographic data for two open TTE datasets, TMED-2 and MIMIC-IV-ECHO.
\item A systematic analysis of the available sociodemographic data of 6 open TTE datasets, finding a lack of descriptions of how such data is collected and limited representation across racial and ethnic groups.
\item An exploratory subgroup analysis of two published AS detection models on the TMED-2 dataset, finding insufficient evidence of subgroup validity for sex, race, and ethnic subgroups.
\end{enumerate}

%% file: tables/new_data.tex
\begin{table*}[ht]
  {
\caption{
Summary of our work's data contributions; \checkmark denotes reporting present in existing data.
We contribute new and improved sociodemographic data for MIMIC-IV-ECHO and TMED-2 respectively. We also report sociodemographic composition for the  predefined train, validation, and test splits for TMED-2, and analyze classifier performance by subgroup.
No previous work has reported machine learning model performance across subgroups on these echocardiogram datasets.
}
\label{tab:new_data}
}
  {\begin{tabular}{llllll}
  \toprule
Dataset & Sex/Gender & Race/Ethnicity & Age & Per Split & Classifier Perf.~by Subgroup \\
  \midrule
TMED-2 & \textbf{Improved} & \textbf{Improved} & \textbf{Improved} & \textbf{New} & \textbf{New} \\
MIMIC-IV-ECHO & \textbf{New} & \textbf{New} & \textbf{New}  \\
EchoNet-Dynamic & \checkmark &  & \checkmark & \checkmark  \\
EchoNet-LVH & \checkmark & \checkmark & \checkmark & \checkmark \\
Unity & \checkmark &  & \checkmark & \checkmark  \\
CAMUS & & & & \\
  \bottomrule
  \end{tabular}}
\end{table*}

%% file: sections/related_work.tex
\section{Related Work}
\label{sec:related_work}

\subsection{Sociodemographic Reporting Guidelines}

The SAGER guidelines \citep{heidari2016sex} advocate for explicit definitions of sex and gender as well as disaggregating the patient population statistics and experiment results across \textbf{both} sex and gender.
The authors acknowledge that guidance on gender-diverse (e.g. transgender, intersex) populations is not provided and encourage researchers to consider the relevance of their work for gender-diverse populations.
Gender-diverse populations are currently underrepresented in cardiovascular research \citep{caceres2017systematic} and within cardiology \citep{disparitiesNotBinary}.

\citet{porter2024reporting} summarize the ways that sex and gender affect cardiovascular health and how the SAGER guidelines can be adapted for cardiovascular research.
Notably, they recommend recording patient sexual orientation, sex, and gender. 

The C4DISC guidelines \citep{2022Race} advocate for explicitly defining the concepts of race, ethnicity, and nationality as well as being as specific as possible when describing a racial, ethnic, or national group.
Throughout this paper we use the phrases ``race/ethnicity'' and ``race and ethnicity'' due to the existing sociodemographic reporting associating a single racial or ethnic group to a person instead of separately considering race and ethnicity.

TRIPOD-AI \citep{collins2024tripod+ai} guidelines for evaluating predictive models advocate for describing known healthcare inequalities and reporting cohort information (e.g. patient counts) for key demographic groups.
The extension TRIPOD-LLM \citep{gallifant2025tripod+llm} removes these points, making how to evaluate bias more vague.

\subsection{Sociodemographics in ML for Healthcare}

Many previous works focus on how model performance across subgroups.
\citet{gianfrancesco2018potential} investigate subgroup bias for electronic health record data, identifying similar challenges for data collection and prediction but proposing solutions focused on monitoring of model outputs.
\citet{yagi2022importance} find that \textit{electro}cardiogram models perform well across subgroups, but varies based on abnormalities caused by cardiovascular disorders.
\citet{movva2023coarse} investigate the differences between model performance disparities calculated on coarse and granular race data, finding that the common use of coarse data may underestimate racial disparities.
Our work puts a larger focus on the sociodemographic data necessary for these analyses and how subgroup analyses relates to subgroup validity.

\citet{jimenez2025picture} provide case studies on four types of medical imaging: chest x-rays, skin lesions, brain MRI, and fetal ultrasound. For each modality, the authors emphasize the medical impact of lacking sociodemographic reporting. They also highlight the importance of documenting study design and patient demographics, ultimately recommending a system of ever-updated \emph{living} documentation for medical imaging data. 

%% file: sections/data.tex
\section{New Sociodemographic Data}
\label{sec:data}

We describe new and improved sociodemographic data for TMED-2 in Sec.~\ref{sec:tmed-data-contrib} and for MIMIC-IV-Echo in Sec.~\ref{sec:mimic-data-contrib}. 
Our overall data contributions are summarized in Table \ref{tab:new_data}.
Per-dataset subgroup composition is summarized in Tables \ref{tab:race_tmed_labels}, \ref{tab:age_sex}, and \ref{tab:race_ethnicity_percent}, with detailed analysis of composition following in Section \ref{sec:analysis}.

\subsection{TMED-2: Improved subgroup data}
\label{sec:tmed-data-contrib}

\input{tables/race_tmed_labels}

TMED-2 \citep{huang2022TMED2,huangNewSemisupervisedLearning2021} is an echocardiogram dataset collected from Tufts Medical Center in Boston, MA, representing patients receiving TTEs during routine care at a major cardiology center.
TMED-2 is focused on enabling automated screening of aortic stenosis (AS), with 599 TTEs with expert-assigned labels for ``No AS,'' ``Early AS,'' and ``Significant AS''.
In addition to this labeled data, unlabeled echocardiograms are provided to enable self and semi-supervised learning methods.
However, we focus our work on the labeled subset of TMED-2.

For TMED-2 we obtained (with IRB approval) sex, race, and ethnicity data for each AS-labeled TTE scan. Previous open releases did not provide this information, and previous publications only provided an abbreviated summary table~\citep{tmedJASE}.
Sex designates administrative sex assigned at birth when available, which should be self-reported but may not be in practice \citep{boehmer2002self, samalik2023discrepancies}.
Race and ethnicity is similarly reported.
We have eliminated the previously reported ``Other'' \citep{tmedJASE} race and ethnicity designation via improved parsing of electronic health record data, finding that this group was in actuality predominately Asian or unspecified.
This highlights how working to eliminate ambiguous designations can produce more accurate sociodemographic reporting.

TMED-2's overall subgroup composition is reported in Tables \ref{tab:age_sex} and \ref{tab:race_ethnicity_percent}, with a breakdown by AS-severity label in Table~\ref{tab:race_tmed_labels}. See Section~\ref{sec:analysis} for detailed discussion and comparison to other datasets.

\subsection{MIMIC-IV-ECHO: New subgroup data}
\label{sec:mimic-data-contrib}

MIMIC-IV-ECHO \citep{mimicivecho} contains echocardiogram imaging from 2017 - 2019 for 4,579 distinct patients; each patient's broader electronic health record is available in the MIMIC-IV Clinical Database~\citep{johnson2023mimiciv}.
Both datasets are available on PhysioNet \citep{goldberger2000physiobank}.
Unlike TMED-2, no labels or predefined machine learning task is associated with MIMIC-IV-ECHO, though diagnostic codes or text records in the clinical database could be used to determine labels.

We report sex, race, and ethnicity for MIMIC-IV-ECHO by associating each echocardiogram with corresponding sociodemographic data in the MIMIC-IV clinical data.
According to the data documentation \citep{mimic-iv-docs}, sex is ``genotypical sex,'' despite no mention of how genotyping was performed. 
Race and ethnicity is separately documented for each hospital admission, which can lead to inconsistent reporting.
We resolve this issue by using each patient's most frequently reported race and ethnicity.

MIMIC-IV-ECHO's overall subgroup composition is reported in Tables \ref{tab:age_sex} and \ref{tab:race_ethnicity_percent}. 
In addition to the coarse race/ethnicity labels there, we are able to use detailed race and ethnicity data in MIMIC-IV.
Following previous work \citep{movva2023coarse}, we report finer-grained race and ethnicity categories in Table \ref{tab:mimic_granular}, such as ``Chinese'' instead of ``Asian.'' 
This table also clarifies that the ``Other'' category we report is taken directly from the electronic health record with the exception of a single patient that reported more than one race and ethnicity.

%% file: tables/race_tmed_labels.tex
\begin{table*}[htbp]
  {\caption{Percentage and count of AS severity labels for TMED-2 scans by race and sex.
  Percentages sum to 100\% across rows.
White patients have smaller ``No AS'' percentage than other racial/ethnic groups.
}
\label{tab:race_tmed_labels}
}
{
\begin{tabular}{l r r r r r r r r r r}
\toprule
 & Total && \multicolumn{2}{c}{No AS} 
 &~& \multicolumn{2}{c}{Early AS}
 &~& \multicolumn{2}{c}{Significant AS} 
 \\
\midrule
White & 505 && 18.4\% & 93 && 27.3\% & 138 && 54.3\% & 274 \\
Black & 27 && 40.7\% & 11 && 25.9\% & 7 && 33.3\% & 9 \\
Asian & 43 && 34.9\% & 15 && 39.5\% & 17 && 25.6\% & 11 \\
Latino & 17 && 47.1\% & 8 && 11.8\% & 2 && 41.2\% & 7 \\
Unspecified & 7 && 0.0\% & 0 && 100.0\% & 7 && 0.0\% & 0 \\
\midrule
Female & 254 && 20.1\% & 51 && 29.5\% & 75 && 50.4\% & 128 \\
Male & 338 && 22.5\% & 76 && 26.3\% & 89 && 51.2\% & 173 \\
Unspecified & 7 
&& 0.0\% & 0 && 100.0\% & 7 && 0.0\% & 0 
\\
\bottomrule
\end{tabular}
}
\end{table*}

%% file: sections/analysis.tex
\section{Subgroup Composition Analysis of 6 TTE Datasets}
\label{sec:analysis}

\input{tables/age_sex}

Here, we contribute a critical analysis of the subgroup composition of 6 open TTE datasets. 
We cover the new data for TMED-2 and MIMIC-IV-ECHO from Section~\ref{sec:data} as well as off-the-shelf sociodemographics from 4 additional open datasets.
In Section \ref{sec:other_datasets}, we describe the 4 additional datasets, especially  their handling of sociodemographic reporting.
Later, we identify trends in sociodemographics across datasets and the shortcomings of current reporting practices, focused on 
sex and gender in Section \ref{sec:meta_sex} and race and ethnicity in Section \ref{sec:meta_race}.
In Section \ref{sec:subgroup_labels} we describe trends in aortic stenosis labels across subgroups.

\subsection{Additional Datasets Analyzed}
\label{sec:other_datasets}

\textbf{EchoNet-LVH} \citep{echonet-lvh} was collected for automating measurements of the left ventricle.
It is the only existing echocardiogram dataset that reports age, sex, and race and ethnicity.
It is composed of four cohorts, distinguished by the hospital name where the data was collected, Stanford Health Care (SHC) in Stanford, CA or Cedars-Sinai Medical Center (CSMC) in Los Angeles, CA, and the echocardiography view-type, PLAX or A4C.

\textbf{EchoNet-Dynamic} \citep{echonet-dyn} was collected for automating left ventricle image segmentation, ejection fraction estimation, and  cardiomyopathy assessment.
\textbf{Unity} \citep{unity} was collected for automated labeling of key points in echocardiograms, with these points enabling measurements such as interventricular septum thickness.
Both datasets report age and Unity reports sex while EchoNet-Dynamic is ambiguous about whether it is reporting sex or gender.
Neither dataset reports race and ethnicity. 

\textbf{CAMUS} \citep{camus} was collected for automating the measurement of left ventricular volumes and ejection fraction.
No sociodemographic data was reported for it.

The discussion of sociodemographic data in these works is extremely limited, with the only case being a sentence in the supplemental reporting summary of EchoNet-Dynamic that notes the demographics are consistent with those of the entire patient population undergoing imaging at that hospital.
There is no discussion of how the sociodemographic data was collected, how sociodemographics may affect the evaluation of model performance, or how models could impact subgroups of patients upon deployment.

\subsection{Sex and Gender Composition}
\label{sec:meta_sex}

Table \ref{tab:age_sex} reports the number of patients of each sex for each echocardiogram dataset.
There is an even split between male and female patients for most datasets, but TMED-2 and the EchoNet-LVH data from Cedars-Sinai Medical Center both have a 15 percentage points or more skew towards male patients.

Gender-diverse populations are currently excluded from the data due to reporting reflecting a binary view of sex and no consideration of gender.
In the cases of TMED-2 and MIMIC-IV-ECHO, this binary encoding of sex is all that was available in the electronic health record.
Recording patient sex, gender, and sexual orientation via methods similar to those proposed by \citet{porter2024reporting} is necessary to enable the study of and provide better care for gender-diverse patients.

\subsection{Race and Ethnicity Composition}
\label{sec:meta_race}
\input{tables/race_ethnicity_percent}

Table \ref{tab:race_ethnicity_percent} reports the race and ethnicity percentages (counts in Table \ref{tab:race_ethnicity_count}) for each echocardiogram dataset where such data is available.
White patients make up the largest part of each dataset, always making up no less than $40\%$ of the data.
In contrast, other racial and ethnic groups never exceed $20\%$ of the data and frequently make up less than $10\%$.
American Indian and Pacific Islander patients are the least represented, always at less than $2\%$ of the data.

The ``Other'' category for EchoNet-LVH is described as ``Other racial and ethnic groups were not available because they were not included in the electronic health records'' \citep{echonet-lvh}. Those authors likely encountered ``Other'' as a category in their raw data as we did with MIMIC-IV-ECHO.
Future electronic health records will need to allow more detailed descriptions of race and ethnicity, ideally allowing patients to self-describe in their own words, to increase the visibility of these patients in datasets.

Current sociodemographic data also reflects a limited view of race and ethnicity by requiring patients to identify as a single race or ethnicity.
Collapsing race and ethnicity into a single category in this way prevents accurate representation of patients that are multiracial or have more than one racial and ethnic identities.
Data on race and ethnicity should instead allow for patients to have multiple racial and ethnic identities, such as how reporting is handled in the 2020 US Census \citep{USCensus}.

\subsection{AS Labels by Subgroup}
\label{sec:subgroup_labels}

Of all 6 datasets, only TMED-2 provides specific labels for aortic stenosis (AS), a common valve disease. Table \ref{tab:race_tmed_labels} summarizes for TMED-2 patients how AS severity labels are distributed across subgroups for both race and sex.
We find that white patients have significantly less ``No AS'' labels than other racial and ethnic groups.
There is no significant difference in label frequency across sex, with both label distributions being very similar to the white distribution.

%% file: tables/age_sex.tex
\begin{table*}[ht]
  {\caption{
Mean age, sex percentage, and sex patient count in echocardiogram datasets.
TMED-2 and EchoNet-LVH data from CSMC underrepresent female patients.
The age reported for Unity is the median as the mean is not reported.
Abbreviations of EchoNet cohorts are in caption of Tab.~\ref{tab:race_ethnicity_percent}
}
\label{tab:age_sex}
}
{\begin{tabular}{lrrrc}
  \toprule
Dataset & \multicolumn{1}{l}{Age Mean} & \multicolumn{1}{l}{Male} & \multicolumn{1}{l}{Female} & \multicolumn{1}{l}{Unspecified} \\
\midrule
TMED-2 & 70.9 & 56.85\% (328) & 41.94\%	(242) & 1.21\% (7)\\
MIMIC-IV-ECHO & 65.0 & 47.56\% (2178) & 52.44\% (2401) &  \\
\midrule
EchoNet-LVH (SHC:PLAX) & 61.6 & 45.76\% (5492) & 54.24\% (6509) &  \\
EchoNet-LVH (SHC:A4C) & 69.1 & 48.03\% (3883) & 51.97\% (4201) &  \\
EchoNet-LVH (CSMC:PLAX) & 62.8 & 61.73\% (808) & 38.27\% (501) &  \\
EchoNet-LVH (CSMC:A4C) & 69.6 & 68.91\% (1620) & 31.09\% (731) &  \\
EchoNet-Dynamic & 68 & 51.30\% (5145) & 48.70\% (4885) &  \\
Unity & 62* & 41.13\% (928) & 46.63\% (1052) & 12.23\% (276) \\
\bottomrule
\end{tabular}}
\end{table*}

%% file: tables/race_ethnicity_percent.tex
\begin{table*}[htbp]
  {\caption{
Race and ethnicity percentage in echocardiogram datasets.
White patients are far more frequent than other groups.
Abbreviations: AI: American Indian; PI: Pacific Islander; EN-LVH: Echo-Net LVH; SHC: Stanford Health; CSMC: Cedars-Sinai Medical Center; P: PLAX; A: A4C (see Sec.~\ref{sec:other_datasets}).
}
\label{tab:race_ethnicity_percent}
 }
{\begin{tabular}{lllllllll}
\toprule
Dataset & White & Black & Hispanic & Asian & AI & PI & Other & Unspecified \\
\midrule
TMED-2 & 84.58\% & 4.16\% & 2.95\% & 7.11\% & \multicolumn{1}{c}{} & \multicolumn{1}{c}{} & \multicolumn{1}{c}{} & 1.21\% \\
MIMIC-IV-ECHO & 65.78\% & 16.64\% & 3.91\% & 2.51\% & 0.09\% & \multicolumn{1}{c}{} & 2.62\% & 8.45\% \\
\midrule
EN-LVH (SHC:P) & 42.30\% & 3.97\% & 11.07\% & 14.44\% & 0.27\% & 1.47\% & 7.80\% & 18.68\% \\
EN-LVH (SHC:A) & 44.88\% & 3.87\% & 8.24\% & 8.86\% & 0.16\% & 0.83\% & 10.49\% & 22.67\% \\
EN-LVH (CSMC:P) & 53.25\% & 18.26\% & 13.60\% & 6.49\% & 0.31\% & 0.31\% & 6.34\% & 1.45\% \\
EN-LVH (CSMC:A) & 62.44\% & 15.01\% & 11.31\% & 4.47\% & 0.17\% & 0.13\% & 6.21\% & 0.26\%\\
 \bottomrule
\end{tabular}}
\end{table*}

%% file: sections/experiments.tex
\section{Classifier Performance by Subgroup}
\label{sec:results}

\input{tables/sex_bacc_test}

\input{tables/race_bacc_test}

\input{tables/all_auroc_test}

We now examine classifier performance by subgroup, focused on the AS disease classification task
for TMED-2 data.

\subsection{Experiment Design and Protocol}

We examine two separately published models for AS classification.
First, SAMIL~\citep{huangDetectingHeartDisease2023} is a model trained specifically for AS on TMED-2 data.
Second, PanEcho~\citep{holsteCompleteAIEnabledEchocardiography2025} is a multi-task model developed on closed-source TTE data from Yale New Haven Health in New Haven, CT.
AS classification is one of its 39 tasks; \citeauthor{holsteCompleteAIEnabledEchocardiography2025} report strong performance, AUROC $\geq$ 0.98, for severe AS detection on a closed-source external validation set from Semmelweis University in Budapest, Hungary.
Subgroup \emph{differences} across sex (male/female) and race (white/Black/``other'') for PanEcho were reported, but raw metrics per subgroup were not provided.

TMED-2 provides 5 possible fine-grained AS grades (None, Mild, Mild-to-Moderate, Moderate, and Severe) for each patient-scan from a board-certified expert following guidelines~\citep{baumgartnerRecommendationsEchocardiographicAssessment2017}. 
Motivated by different goals, each published model was trained using a different coarsening of fine-grained levels into a simpler 3-way classification task.
Each model keeps ``None'' as its own class, but has different ways of grouping the rest detailed in Appendix~\ref{apd:AS_class_labels}.

As a primary common evaluation, we evaluate each model's ability to distinguish between ``No AS'' and ``Some AS'', where ``Some'' includes Mild, Mild-to-Moderate, Moderate, and Severe grades. 
This binary task represents a clinically-useful deployment scenario~\citep{tmedJASE}.
Performance at this binary task, measured via area under the ROC curve (AUROC), can be compared across both SAMIL and PanEcho.
Additionally, we report  balanced accuracy on the distinct 3-way classification tasks for each model.
Balanced accuracy results are not comparable across SAMIL and PanEcho due to different class mappings.
Both metrics are computed for the 3 predefined dataset splits of TMED-2~\citep{huang2022TMED2} then aggregated by taking the average and 95 percentile confidence interval of 3,000 bootstrap samples (1,000 for each split) \citep{acc_comparison}.



Both SAMIL and PanEcho take as input many image and video instances gathered in a routine TTE scan and make one patient-scan prediction of AS severity. 
One patient-scan contains several instances of 2D images/videos that show a spatial cross-section of the heart from different canonical viewpoints, Doppler images/videos that depict blood flow over time, and other types of data not used by either model.
The input instances used by each model for prediction differ to best match how each was trained.
SAMIL's input is only 2D instances, keeping still images and the first frame only of any videos at 112x112 resolution.
SAMIL uses a trained attention mechanism to prioritize some instances over others in its prediction.
In contrast, PanEcho's input is all 2D and Doppler instances, using only videos at least 16 frames long.
Given a 256x256 video, PanEcho processes four separate clips of 16 consecutive frames, each starting at random indices, averaging predictions from each clip for a per-video prediction..
An unweighted average across videos produces the patient-scan prediction.
Preprocessing and implementation details are fully described in Appendix~\ref{apd:AS_classifier_details}.


\subsection{Analysis by Subgroup}

\textbf{Sex.}
We report for both models the AUROC by sex subgroup in \ref{tab:all_auroc_test}, with balanced accuracy for each model in Tables \ref{tab:sex_bacc_test} and \ref{tab:sex_bacc_panecho_test}.

Average performance across models and metrics slightly favor female patients, although the bootstrapped difference indicates that the difference is not statistically significant with 95\% confidence.
Such a result would be unintuitive because models  typically perform better on subgroups with more data than on subgroups with less data \citep{buolamwini2018gender, doi:10.1073/pnas.1919012117} and Table \ref{tab:race_tmed_labels} shows little difference in disease prevalence.
It remains an open question whether future experiments with more data or more consistent model performance across dataset splits will eliminate or exacerbate this trend.

\textbf{Race and Ethnicity.}
We report the AUROC by race and ethnicity for both models in Table~\ref{tab:all_auroc_test}, with balanced accuracy results for each model in Tables \ref{tab:race_bacc_test} and \ref{tab:race_bacc_panecho_test}.
The results for SAMIL at first glance appear in line with expectations, with better performance on white patients than the less prevalent Black or Asian subgroups.
However, the small sample sizes for subgroups other than white patients prevents these results from being statistically significant.
This disparity is clearly seen in the confidence intervals for balanced accuracy, which cover $21.1$ percentage points for white patients and between $66.6$ and $100.0$ percentage points for other subgroups.

These differences in confidence intervals highlight the impact of small sample sizes on these subgroup analyses.
Our results are exploratory primarily due to TMED-2's small sample sizes preventing the subgroup analysis from being statistically significant, even before accounting for multiple testing with the Benjamini–Hochberg procedure \citep{benjamini1995controlling}.
Future evaluations of echocardiogram models on other datasets may also suffer from small per-subgroup sample sizes given broader trends in TTE datasets from Section~\ref{sec:analysis}.

%% file: tables/sex_bacc_test.tex
\begin{table*}[htbp]
  {\caption{
Balanced accuracy (3-class) by sex for SAMIL model on on TMED-2's 3 predefined test sets, with scan count in parentheses.
}
\label{tab:sex_bacc_test}
}
{\begin{tabular}{llll}
\toprule
 & Female (254) & Male (338) & Unspecified (7) \\
\midrule
SAMIL 0 & 89.4 (53) & 60.1 (67) & — (0) \\
SAMIL 1 & 71.3 (54) & 75.6 (65) & 100.0 (1) \\
SAMIL 2 & 80.9 (46) & 73.8 (74) & — (0) \\
\midrule
SAMIL Av. & 80.5 & 69.8 & 100.0 \\
SAMIL CI & [62.0, 94.2] & [50.3, 83.9] & — \\
\bottomrule
\end{tabular}}
\end{table*}

%% file: tables/race_bacc_test.tex
\begin{table*}[htbp]
  {\caption{
Balanced accuracy (3-class) by race for SAMIL model on on TMED-2's 3 predefined test sets, with scan count in parentheses.
}
\label{tab:race_bacc_test}
}
{\begin{tabular}{llllll}
\toprule
 & White (505) & Black (27) & Asian (43) & Latino (17) & Unspecified (7) \\
\midrule
SAMIL 0 & 73.8 (105) & 33.3 (5) & 62.5 (6) & 100.0 (4) & — (0) \\
SAMIL 1 & 70.5 (100) & 83.3 (4) & 58.3 (7) & 100.0 (8) & 100.0 (1)\\
SAMIL 2 & 76.8 (106) & 66.7 (4) & 44.4 (7) & 66.7 (3) & — (0)\\
\midrule
SAMIL Av. & 73.6 & 62.7 & 59.7 & 88.7 & 100.0\\
SAMIL CI & [62.7, 83.8] & [0.0, 100.0] & [33.3, 100.0] & [33.3, 100.0] & — \\
\bottomrule
\end{tabular}}
\end{table*}

%% file: tables/all_auroc_test.tex
\begin{table*}[htbp]
  {\caption{
AUROC (2-class) for different models on TMED-2's 3 predefined test sets by subgroups and with scan count in parentheses.
Unspecified is dropped due to not having enough patients with varying labels to calculate.
}
\label{tab:all_auroc_test}
}
{\begin{tabular}{lllll|ll}
\toprule
 & White (505) & Black (27) & Asian (43) & Latino (17) & Female (254) & Male (338) \\
\midrule
SAMIL 0 & 0.93 (105) & 0.50 (5) & 1.00 (6) & 1.00 (4) & 0.99 (53) & 0.88 (67) \\
SAMIL 1 & 0.93 (100) & 1.00 (4) & 1.00 (7) & 1.00 (8) & 0.95 (54) & 0.95 (65) \\
SAMIL 2 & 0.98 (106) & 0.50 (4) & 0.75 (7) & — (3) & 0.99 (46) & 0.95 (74) \\
\midrule
SAMIL Av. & 0.95 & 0.52 & 0.87 & 0.61 & 0.98 & 0.93 \\
SAMIL CI & [0.87, 1.00] & [0.00, 1.00] & [0.00, 1.00] & [0.00, 1.00] & [0.89, 1.00] & [0.80, 0.99] \\
\midrule
\midrule
PanEcho 0 & 0.93 (105) & 1.00 (5) & 1.00 (6) & 1.00 (4) & 1.00 (53) & 0.93 (67)\\
PanEcho 1 & 0.81 (100) & 1.00 (4) & 1.00 (6) & 1.00 (8) & 0.93 (54) & 0.87 (65)\\
PanEcho 2 & 0.93 (106) & 1.00 (4) & 1.00 (7) & 0.00 (3) & 0.99 (46) & 0.92 (74)\\
\midrule
PanEcho Av. & 0.89 & 0.82 & 0.95 & 0.61 & 0.97 & 0.91\\
PanEcho CI & [0.71, 0.98] & [0.00, 1.00] & [0.00, 1.00] & [0.00, 1.00] & [0.81, 1.00] & [0.77, 0.98]\\
\bottomrule
\end{tabular}}
\end{table*}

%% file: sections/conclusion.tex
\section{Conclusion}
\label{sec:conclusion}

Our work has highlighted that subgroup analyses are challenging even when following the current best practices for sociodemographic reporting.
These challenges necessitate changes in dataset creation, sociodemographic reporting, and model analysis.

We propose three changes to typical machine learning for healthcare practices that are necessary to better evaluate subgroup validity in machine learning for echocardiogram data.
First, more data is necessary for underrepresented subgroups.
This is especially necessary for traditionally underrepresented racial and ethnic groups, such as American Indian and Pacific Islander patients.
Future work may proactively recruit these subgroups~\citep{romanBridgingUnitedStates2025}, as random sampling of the patient population would likely provide little additional data. Proactive work is especially needed if source hospitals remain mostly from the U.S. or Europe to avoid amero- or euro-centric trends in vision data~\citep{shankarNoClassificationRepresentation2017}.

Second, researchers should improve the specificity of sociodemographic reporting beyond what is typical in electronic health records.
This is necessary to gather data on subgroups not currently represented in data collection, such as gender-diverse or multiracial patients.
Better specificity could replace an overly-broad ``Asian'' category to   distinguish groups like South Asians who are known to have higher risk of cardiovascular disease~\citep{south_asian_rep}.
It would also benefit patients currently assigned to the ambiguous ``Other'' racial and ethnic category.

Finally, subgroup-focused validity analyses need to be performed on predictive models before deployment.
Our results show that subgroup validity concerns can only be addressed by further analyses that benefit from larger, more representative datasets.
Furthermore, reproducibility remains a concern for ML overall~\citep{mcdermottReproducibilityMachineLearning2021}, and the reproducibility of subgroup analyses is necessary to ensure credibility for care decisions upon model deployment.

%% file: sections/appendix.tex
\section{Race and Ethnicity Patient Counts}
\label{apd:race_count}

\input{tables/race_ethnicity_count}
\input{tables/mimic_granular}

Table \ref{tab:race_ethnicity_count} reports the patient counts for race and ethnicity across datasets.
This table was separated from the percentages for better readability, not for any difference in calculation compared to Table \ref{tab:age_sex}, which contains both.

Table \ref{tab:mimic_granular} reports fine-grained race and ethnicity categories using the same electronic health record parsing as \citet{movva2023coarse}.
We are reporting only on the MIMIC-IV-ECHO patients with sociodemographic data in MIMIC-IV, instead of the sociodemographics of all of MIMIC-IV as done in previous work.

\newpage
\section{Classifier Implementation Details}
\label{apd:AS_classifier_details}

\subsection{Class Labels for AS Severity}
\label{apd:AS_class_labels}
Within TMED-2, each patient-scan is assigned one of 5 fine-grained AS severity levels, listed below in Table~\ref{tab:label_mapping}. Motivated by distinct goals, the creators of SAMIL~\citep{huangDetectingHeartDisease2023} and PanEcho~\citep{holsteCompleteAIEnabledEchocardiography2025} intend their models for different 3-class mappings of AS severity. Table~\ref{tab:label_mapping} below depicts how we mapped the 3-class systems of each published classifier to the fine-grained grades available in TMED-2.

\begin{table}[!h]
    \begin{tabular}{l | c | c | c | c | c |}
    \toprule
    TMED-2's fine-grained AS grades & No AS & Mild AS & Mild-to-Moderate & Moderate AS & Severe AS
    \\ \midrule
    SAMIL's 3 classes & No AS & \multicolumn{2}{c|}{Early AS} & \multicolumn{2}{c|}{Significant AS}
    \\
    PanEcho's 3 classes & No AS & \multicolumn{3}{c|}{Mild-Moderate AS} & Severe AS \\
    \bottomrule
    \end{tabular}
    \caption{Mapping of labels from each classifier to fine-grained labels in TMED-2.}
    \label{tab:label_mapping}
\end{table}



SAMIL~\citep{huangDetectingHeartDisease2023} keeps "None" unchanged, groups Mild together with Mild-to-Moderate as “Early,” and Moderate together with Severe as “Significant”. The balanced accuracy results for SAMIL assess this None/Early/Significant 3-class task.

The PanEcho model~\citep{holsteCompleteAIEnabledEchocardiography2025} groups Mild, Mild-to-Moderate, and Moderate collectively as “Mild-Moderate,” while leaving None and Severe unchanged. The balanced accuracy results for PanEcho assess this None/Mild-Moderate/Severe 3-class task. 

\subsection{Performance Metric Computation and Interpretation}

There are 3 predefined train/validation/test splits of TMED-2 \citep{huang2022TMED2}; each one was created by a different random partition and has roughly 360/120/120 patients in the train/validation/test sets. Each patient's data belongs entirely to one of training, validation, or test sets. To overcome the limited information in a single test set, for each model, we evaluate separately on all 3 test sets and report the average and 95\% confidence interval.

For PanEcho, we used pretrained weights found on that project's github \url{https://github.com/CarDS-Yale/PanEcho} in August 2025. The same model made predictions on each of the 3 TMED-2 test sets.

SAMIL is trained on TMED-2 data. For this work, we reused code from \citet{huangDetectingHeartDisease2023} but retrained models from scratch, as the exact models reported on in \citet{huangDetectingHeartDisease2023} were no longer available.
For each split, we train on that split's train set and perform hyperparameter tuning via grid search on that split's validation set. Thus, performance reported for SAMIL represents a common architecture but separate neural network weights for each split.

\textbf{Balanced accuracy} is computed as an unweighted average across classes, where for each class we count the fraction of its true members that were classified correctly. This metric 
Balanced accuracy metrics are not comparable across the two models, because the 3-way class definitions are different (see Table~\ref{tab:label_mapping}). 

\textbf{AUROC} results for the No-vs-Some AS binary task \emph{are} comparable across the two models, as that class mapping is consistent for the two models.



\subsection{Selection and preprocessing of input instances for each model}

At a high level, TMED-2 contains a collection of data modalities that are common to any routine echocardiogram: grayscale 2D videos and still frames that represent a spatial cross-section of the heart's complex three-dimensional anatomy; Doppler videos and still frames that depict bloodflow over time, and color-masked 2D videos and still frames that depict spatial cross-sections with machine-annotated colors indicating blood movement.

\paragraph{SAMIL input.}
The SAMIL model, following~\citet{huangDetectingHeartDisease2023}, does not incorporate Doppler instances or color-masked 2D instances. Thus, to evaluate SAMIL in the same way it was trained, we excluded those modalities from SAMIL’s inputs during TMED-2 evaluation. It is important to note that this exclusion did not remove any studies from evaluation.
For every eligible video/image, we took the first frame, downsampled each selected image to 112 × 112 with three channels, and finally assembled a bag of F images with shape F x 3 × 112 × 112, where F varies per study. This was then provided as input and the corresponding study-level predictions were recorded. The final trained SAMIL model was selected after hyperparameter tuning on a held out validation dataset for variables such as learning rate, weight decay, etc.

\paragraph{PanEcho input.}
PanEcho required a different preprocessing pipeline. We attempted to follow the methodology as described in the original PanEcho paper \citep{holsteCompleteAIEnabledEchocardiography2025} and the accompanying codebase as closely as possible.  When needed we attempted to document and justify any assumptions made. At a high-level, PanEcho consumes variable-length echocardiogram videos and Doppler videos. The PanEcho's prediction routine creates a fixed number of clips per video (set to 4 in the released code), where each clip has a set length (16 frames in the released code). PanEcho returns one probabilistic classification per video, via an unweighted average over predictions from each clip. 
Once video-level predictions have been collected, we compute an unweighted average across all videos in an individual patient-scan to get the scan-level prediction. 

For input instance selection, following the description in \citet{holsteCompleteAIEnabledEchocardiography2025} we provided grayscale 2D instances, color-masked 2D instances, and Doppler instances as input. Our reading of the paper is that training was conducted utilizing solely video inputs. We therefore ran experiments that included Doppler videos versus removing them, and single-frame stills versus removing them. We found that utilizing solely video inputs (not still frames) and including both 2D and Doppler videos gave the best performance. 
We therefore excluded any 2D or Doppler instance that did not contain at least 16 frames.


Preprocessing input instances from TMED-2 for PanEcho required new code to be written as the details described in \citeauthor{holsteCompleteAIEnabledEchocardiography2025}'s supplementary materials were not fully implemented in the public codebase. Their original paper states that videos masked out pixels beyond the central image content (the canonical ``piece-of-a-donut''-shaped viewing window of a 2D ultrasound image) via a fixed-threshold binarization, followed by masking all pixels outside the convex hull of the largest contour. Frames were then cropped to contain just the central content in a temporally consistent manner and then downsampled to 256 × 256 with bicubic interpolation. Because this exact pipeline was not available, we implemented our best approximation of the masking and recentering procedure. Our provided code provides this preprocessing logic.

\textbf{Hyperparameters.}
For hyperparameters, we defaulted to the values outlined in \citeauthor{holsteCompleteAIEnabledEchocardiography2025}'s paper and codebase. By this logic we set the number of clip permutations per video to 4 and the clip length to 16 frames. 
Two key hyperparameters were ambiguous: namely normalization and masking. First, it was unclear whether video inputs should be normalized using the mean and standard deviation of pixel values from ImageNet (line 44 of their \href{https://github.com/CarDS-Yale/PanEcho/blob/836c46ef6bc4eace50b7dc36704ecef2d33affde/src/dataset.py#L44}{dataset.py} suggests this normalization; comments in their scripts also suggest PanEcho's architecture was pretrained on ImageNet). Second, because preprocessing was constructed in-house to approximate the described masking and cropping, we wished to verify performance with and without this preprocessing. Thus, we conducted four additional ablations that crossed normalization choice (ImageNet statistics versus no normalization) with masking choice (our masking pipeline enabled versus disabled). We ultimately report results using both the mask-based preprocessing and ImageNet normalization, as this setting yielded the strongest and most consistent performance across the No-vs-Some binary task, as measured by AUROC, and PanEcho's 3-class task, as measured by balanced accuracy. These hyperparameter configurations were chosen based on each model's performance on the TMED-2 validation set. For the combination that performed the best, we report its test set performance.


\section{Classifier Performance on Train and Validation Sets}
\label{apd:results}

\input{tables/sex_bacc_train_val}
\input{tables/sex_auroc_train_val}
\input{tables/race_bacc_train_val}
\input{tables/race_auroc_train_val}
\input{tables/race_bacc_panecho_test}
\input{tables/sex_bacc_panecho_test}

Tables \ref{tab:sex_bacc_train_val} and \ref{tab:sex_auroc_train_val} report the results on sex subgroups for the train and validation sets.
We find lower differences in metrics across sex and smaller confidence intervals due to better model fit.

Tables \ref{tab:race_bacc_train_val} and \ref{tab:race_auroc_train_val} report the results on racial and ethnic subgroups for the train and validation sets.
These validation set metrics are more similar to the test set metrics, likely due to the small subgroup sizes limiting the impact of those subgroups on the hyperparameter tuning process.

\section{PanEcho Balanced Accuracy}

Tables \ref{tab:race_bacc_panecho_test} and \ref{tab:sex_bacc_panecho_test} report the results of PanEcho for both race and ethnicity and sex respectively.
These results are worse than the corresponding AUROC results would suggest.
PanEcho places too little probability on ``Mild-Moderate AS'' and ``Severe AS'' in the ternary task.
However, the summed ``Some AS'' probability used in the binary task compensates for this.

%% file: tables/race_ethnicity_count.tex
\begin{table*}[htbp]
  {\caption{
Race and ethnicity patient count in echocardiogram datasets.
White patients are far more frequent than any other group.
For brevity, American Indian (AI) and Pacific Islander (PI) have been abbreviated.
}
\label{tab:race_ethnicity_count}
}
{\begin{tabular}{lllllllll}
\toprule
Dataset & White & Black & Hispanic & Asian & AI & PI & Other & Unspecified \\
\midrule
TMED-2 & 488 & 24 & 17 & 41 & \multicolumn{1}{c}{} & \multicolumn{1}{c}{} & \multicolumn{1}{c}{} & 7 \\
MIMIC-IV-ECHO & 3012 & 762 & 179 & 115 & 4 & \multicolumn{1}{c}{} & 120 & 387 \\
EchoNet-LVH (SHC:PLAX) & 5077 & 476 & 1328 & 1733 & 32 & 177 & 936 & 2242 \\
EchoNet-LVH (SHC:A4C) & 3628 & 313 & 666 & 716 & 13 & 67 & 848 & 1833 \\
EchoNet-LVH (CSMC:PLAX) & 697 & 239 & 178 & 85 & 4 & 4 & 83 & 19 \\
EchoNet-LVH (CSMC:A4C) & 1468 & 353 & 266 & 105 & 4 & 3 & 146 & 6\\
 \bottomrule
\end{tabular}}
\end{table*}

%% file: tables/mimic_granular.tex
\begin{table*}[htbp]
  {\caption{
Fine-grained race and ethnicity percentages and counts for MIMIC-IV-ECHO.
}
\label{tab:mimic_granular}
}
  {\begin{tabular}{ll}
  \toprule
White & 62.79\% (2669) \\
Black - African American & 15.53\% (660) \\
White - Russian & 3.58\% (152) \\
White - Other European & 3.53\% (150) \\
Other & 2.80\% (119) \\
Black - Cape Verdean & 1.62\% (69) \\
Hispanic/Latino - Puerto Rican & 1.60\% (68) \\
Asian - Chinese & 1.55\% (66) \\
Unknown & 0.94\% (40) \\
Hispanic/Latino & 0.92\% (39) \\
Hispanic/Latino - Dominican & 0.75\% (32) \\
Asian & 0.71\% (30) \\
White - Eastern European & 0.66\% (28) \\
Black - Caribbean Island & 0.42\% (18) \\
Patient Declined to Answer & 0.40\% (17) \\
Black - African & 0.35\% (15) \\
Asian - South East Asian & 0.24\% (10) \\
Hispanic/Latino - Salvadoran & 0.21\% (9) \\
White - Portuguese & 0.21\% (9) \\
Asian - Asian Indian & 0.19\% (8) \\
Hispanic/Latino - Guatemalan & 0.19\% (8) \\
Hispanic/Latino - South American & 0.16\% (7) \\
Hispanic/Latino - Cuban & 0.14\% (6) \\
Hispanic/Latino - Columbian & 0.14\% (6) \\
White - Brazilian & 0.09\% (4) \\
American Indian/Alaska Native & 0.09\% (4) \\
Hispanic/Latino - Central American & 0.05\% (2) \\
Unable to Obtain & 0.05\% (2) \\
Hispanic/Latino - Mexican & 0.02\% (1) \\
Asian - Korean & 0.02\% (1) \\
Multiple Race/Ethnicity & 0.02\% (1) \\
Hispanic/Latino - Honduran & 0.02\% (1) \\
  \bottomrule
  \end{tabular}}
\end{table*}

%% file: tables/sex_bacc_train_val.tex
\begin{table*}[htbp]
  {\caption{
Balanced accuracy on the train and validation sets by sex, with scan count in parentheses.
}
\label{tab:sex_bacc_train_val}
}
{\begin{tabular}{lllllll}
\toprule
 & \multicolumn{2}{c}{Female (254)} & \multicolumn{2}{c}{Male (338)} & \multicolumn{2}{c}{Unspecified (7)} \\
 & Train & Validation & Train & Validation & Train & Validation \\
\midrule
SAMIL 0 & 100.0 (151) & 68.0 (50) & 100.0 (203) & 79.7 (68) & 100.0 (6) & 100.0 (1) \\
SAMIL 1 & 96.7 (149) & 75.0 (51) & 97.9 (207) & 71.0 (66) & 75.0 (4) & 100.0 (2) \\
SAMIL 2 & 98.1 (165) & 76.1 (43) & 97.7 (190) & 71.8 (74) & 100.0 (5) & 50.0 (2) \\
\midrule
SAMIL Average & 98.3 & 73 & 98.5 & 74.2 & 91.7 & 83.3 \\
SAMIL CI & [94.5, 100.0] & [57.3, 85.9] & [95.7, 100.0] & [60.4, 86.5] & [50.0, 100.0] & [0.0, 100.0] \\
\midrule
\midrule
PanEcho 0 & 36.8 (151) & 35.3 (50) & 35.8 (203) & 35.9 (68) & 0.0 (6) & 0.0 (1) \\
PanEcho 1 & 38.4 (149) & 32.6 (51) & 35.8 (207) & 34.4 (66) & 0.0 (4) & 0.0 (2) \\
PanEcho 2 & 35.1 (165) & 37.0 (43) & 36.2 (190) & 36.6 (74) & 0.0 (5) & 0.0 (2) \\
\midrule
PanEcho Average & 36.8 & 35.0 & 36.0 & 35.7 & 0.0 & 0.0 \\
PanEcho CI & [31.9, 41.8] & [25.0, 43.9] & [34.0, 38.5] & [33.3, 39.8] & [0.0, 0.0] & [0.0, 0.0] \\
\bottomrule
\end{tabular}}
\end{table*}

%% file: tables/sex_auroc_train_val.tex
\begin{table*}[htbp]
  {\caption{
AUROC on the train and validation sets by sex, with scan count in parentheses.
}
\label{tab:sex_auroc_train_val}
}
{\begin{tabular}{lllll}
\toprule
 & \multicolumn{2}{c}{Female (254)} & \multicolumn{2}{c}{Male (338)} \\
 & Train & Validation & Train & Validation \\
\midrule
SAMIL 0 & 1.00 (151) & 0.97 (50) & 1.00 (203) & 0.97 (68) \\
SAMIL 1 & 1.00 (149) & 0.91 (51) & 1.00 (207) & 0.94 (66) \\
SAMIL 2 & 1.00 (165) & 0.99 (43) & 1.00 (190) & 0.94 (74) \\
\midrule
SAMIL Average & 1 & 0.96 & 1 & 0.95 \\
SAMIL CI & 	[1.00, 1.00] & [0.81, 1.00] & [1.00, 1.00] & [0.87, 1.00] \\
\midrule
\midrule
PanEcho 0 & 0.89 (151) & 0.83 (50) & 0.87 (203) & 0.89 (68) \\
PanEcho 1 & 0.89 (149) & 0.88 (51) & 0.90 (207) & 0.88 (66) \\
PanEcho 2 & 0.83 (165) & 1.00 (43) & 0.86 (190) & 0.92 (74) \\
\midrule
PanEcho Average & 0.87 & 0.90 & 0.88 & 0.91 \\
PanEcho CI & [0.74, 0.96] & [0.62, 1.00] & [0.80, 0.94] & [0.78, 0.98] \\
\bottomrule
\end{tabular}}
\end{table*}

%% file: tables/race_bacc_train_val.tex
\begin{table*}[htbp]
\tiny
  {\caption{
Balanced accuracy on the train and validation sets by race and ethnicity, with scan count in parentheses.
}
\label{tab:race_bacc_train_val}
}
{\begin{tabular}{lllllllllll}
\toprule
 & \multicolumn{2}{c}{White (505)} & \multicolumn{2}{c}{African Am. (27)} & \multicolumn{2}{c}{Asian (43)} & \multicolumn{2}{c}{Latino (17)} & \multicolumn{2}{c}{Unspecified (7)} \\
 & Train & Valid & Train & Valid & Train & Valid & Train & Valid & Train & Valid \\
 \midrule
SAMIL 0 & 100 (301) & 71.8 (99) & 100 (19) & 75.0 (3) & 100 (24) & 88.9 (13) & 100 (10) & 100 (3) & 100 (6) & 100 (1) \\
SAMIL 1 & 97.4 (304) & 72.9 (101) & 100 (18) & 22.2 (5) & 90.5 (25) & 61.1 (11) & 100 (9) & — (0) & 75.0 (4) & 100 (2) \\
SAMIL 2 & 98.1 (294) & 73.5 (105) & 100 (23) & — (0) & 93.5 (28) & 50.0 (8) & 100 (10) & 50.0 (4) & 100 (5) & 50.0 (2) \\
\midrule
SAMIL Av. & 98.5 & 72.8 & 100 & 48.6 & 94.7 & 66.7 & 100 & 75 & 91.7 & 83.3 \\
SAMIL CI & 	[96.0, 100.0] & [62.9, 81.9] & [100.0, 100.0] & [0.0, 100.0] & [81.0, 100.0] & [33.3, 100.0] & [100.0, 100.0] & [50.0, 100.0] & [50.0, 100.0] & [0.0, 100.0] \\
\midrule
\midrule
PanEcho 0 & 36.6 (301) & 34.6 (99) & 33.3 (19) & 50.0 (3) & 33.3 (24) & 38.9 (13) & 33.3 (10) & 50.0 (3) & 0.0 (6) & 0.0 (1) \\
PanEcho 1 & 36.9 (304) & 33.4 (101) & 33.3 (18) & 50.0 (5) & 35.4 (25) & 33.3 (11) & 33.3 (9) & — (0) & 0.0 (4) & 0.0 (2) \\
PanEcho 2 & 35.8 (294) & 36.4 (105) & 33.3 (23) & — (0) & 35.2 (28) & 33.3 (8) & 33.3 (10) & 50.0 (4) & 0.0 (5) & 0.0 (2) \\
\midrule
PanEcho Av. & 36.4 & 34.8 & 33.3 & 50.0 & 34.6 & 35.2 & 33.3 & 50.0 & 0.0 & 0.0 \\
PanEcho CI & [33.8, 39.1] & [29.5, 39.5] & [33.3, 50.0] & [0.0, 100.0] & [33.3, 52.6] & [33.3, 58.3] & [33.3, 50.0] & [50.0, 100.0] & [0.0, 0.0] & [0.0, 0.0] \\
\bottomrule
\end{tabular}}
\end{table*}

%% file: tables/race_auroc_train_val.tex
\begin{table*}[htbp]
\small
  {\caption{
AUROC on the train and validation sets by race and ethnicity, with scan count in parentheses.
}
\label{tab:race_auroc_train_val}
}
{\begin{tabular}{lllllllllll}
\toprule
\begin{tabular}{lllllllll}
 & \multicolumn{2}{c}{White (505)} & \multicolumn{2}{c}{Black (27)} & \multicolumn{2}{c}{Asian (43)} & \multicolumn{2}{c}{Latino (17)} \\
 & Train & Valid & Train & Valid & Train & Valid & Train & Valid \\
\midrule
SAMIL 0 & 1.00 (301) & 0.96 (99) & 1.00 (19) & 1.00 (3) & 1.00 (24) & 1.00 (13) & 1.00 (10) & 1.00 (3) \\
SAMIL 1 & 1.00 (304) & 0.93 (101) & 1.00 (18) & 0.75 (5) & 1.00 (25) & 0.86 (11) & 1.00 (9) & — (0) \\
SAMIL 2 & 1.00 (294) & 0.95 (105) & 1.00 (23) & — (0) & 1.00 (28) & 1.00 (8) & 1.00 (10) & 1.00 (4) \\
\midrule
SAMIL Av. & 1 & 0.95 & 1 & 0.88 & 1 & 0.95 & 1 & 1 \\
SAMIL CI & [1.00, 1.00] & [0.88, 0.99] & [1.00, 1.00] & [0.00, 1.00] & 	[1.00, 1.00] & [0.60, 1.00] & [1.00, 1.00] & [0.00, 1.00] \\
\midrule
\midrule
PanEcho 0 & 0.86 (301) & 0.79 (99) & 0.93 (19) & 1.00 (3) & 0.87 (24) & 1.00 (13) & 1.00 (10) & 1.00 (3) \\
PanEcho 1 & 0.87 (304) & 0.87 (101) & 0.92 (18) & 1.00 (5) & 0.95 (25) & 0.89 (11) & 1.00 (9) & — (0) \\
PanEcho 2 & 0.81 (294) & 0.95 (105) & 0.94 (23) & — (0) & 0.92 (28) & 0.87 (8) & 1.00 (10) & 1.00 (4) \\
\midrule
PanEcho Av. & 0.85 & 0.87 & 0.93 & 1.0 & 0.91 & 0.92 & 1 & 1 \\
PanEcho CI & [0.75, 0.92] & [0.67, 0.99] & [0.75, 1.00] & [0.00, 1.00] & [0.75, 1.00] & [0.50, 1.00] & [1.00, 1.00] & [0.00, 1.00]
\end{tabular}\\
\bottomrule
\end{tabular}}
\end{table*}

%% file: tables/race_bacc_panecho_test.tex
\begin{table*}[htbp]
  {\caption{
Balanced accuracy (3-class) by race and ethnicity for PanEcho model on TMED-2's 3 predefined test sets, with scan count in parentheses.
}
\label{tab:race_bacc_panecho_test}
}
{\begin{tabular}{llllll}
\toprule
 & White (505) & Black (27) & Asian (43) & Latino (17) & Unspecified (7) \\
\midrule
PanEcho 0 & 36.6 (105) & 33.3 (5) & 50.0 (6) & 33.3 (4) & — (0) \\
PanEcho 1 & 37.3 (100) & 50.0 (4) & 33.3 (7) & 33.3 (8) & 0.0 (1) \\
PanEcho 2 & 37.5 (106) & 33.3 (4) & 33.3 (7) & 0.0 (3) & — (0) \\
\midrule
PanEcho Av. & 37.1 & 38.9 & 38.9 & 22.2 & 0.0 \\
PanEcho CI & [34.0, 41.1] & [0.0, 100.0] & [0.0, 50.0] & [0.0, 50.0] & [0.0, 0.0] \\
\bottomrule
\end{tabular}}
\end{table*}

%% file: tables/sex_bacc_panecho_test.tex
\begin{table*}[htbp]
  {\caption{
Balanced accuracy (3-class) by sex for PanEcho model on TMED-2's 3 predefined test sets, with scan count in parentheses.
}
\label{tab:sex_bacc_panecho_test}
}
{\begin{tabular}{llll}
\toprule
 & Female (254) & Male (338) & Unspecified (7) \\
\midrule
PanEcho 0 & 37.5 (53) & 35.6 (67)  & — (0)  \\
PanEcho 1 & 36.4 (54) & 37.3 (65) & 0.0 (1) \\
PanEcho 2 & 41.1 (46) & 34.3 (74) & — (0) \\
\midrule
PanEcho Av. & 38.3 & 35.7 & 0.0 \\
PanEcho CI & [33.3, 46.7] & [33.3, 40.9] & [0.0, 0.0] \\
\bottomrule
\end{tabular}}
\end{table*}